# Estimating forest carbon stocks from high-resolution remote sensing imagery by reducing domain shift with style transfer

Zhenyu Yu[1, 2], Jinnian Wang[1, 2, *]

**Affiliations:**

[1] School of Geography and Remote Sensing, Guangzhou University, Guangzhou, Guangdong, 510006, China

[2] Innovation Center for Remote Sensing Big Data Intelligent Applications, Guangzhou University, Guangzhou, Guangdong, 510006, China

Corresponding author: Jinnian Wang (yuzhenyu@gzhu.edu.cn)

**Abstract:** Forests function as crucial carbon reservoirs on land, and their carbon sinks can efficiently reduce atmospheric $CO_2$ concentrations and mitigate climate change. Currently, the overall trend for monitoring and assessing forest carbon stocks is to integrate ground monitoring sample data with satellite remote sensing imagery. This style of analysis facilitates large-scale observation. However, these techniques require improvement in accuracy. We used GF-1 WFV and Landsat TM images to analyze Huize County, Qujing City, Yunnan Province in China. Using the style transfer method, we introduced Swin Transformer to extract global features through attention mechanisms, converting the carbon stock estimation into an image translation. We proposed the MSwin-Pix2Pix model, and the results indicated that (1) Swin-Pix2Pix aligned different temporal and spatial distribution through style transfer across domains, which reduced the inter-domain differences caused by sensors, lighting and other factors. Swin-Pix2Pix effectively de-clouded images, and its performance surpassed that of Pix2Pix. (2) In carbon stock estimation, MSwin-Pix2Pix added the median filter module to eliminate anomalous detection using local information. The added mask module effectively excludes non-target areas, thereby reducing model instability. MSwin-Pix2Pix's global feature extraction capability was significantly better than other models (MAE = 16.2891, RMSE = 29.3763, $R^2$ = 0.7105, SSIM=0.7510). (3) In 2005~2020, the total area where carbon stock was 44.04% increased, 10.22% decreased, and 45.74% remained unchanged, indicating an overall increasing trend of carbon stock. It indicated a significant improvement in the ecological environment, laying a good ecological foundation for the county's social and economic development. Our research used the characteristics of the region and forest to achieve a high-resolution carbon stock estimation, providing an important theoretical basis for forest carbon sink regulation.

**Keywords:** carbon stock; forest; style transfer; remote sensing; deep learning

# 1 Introduction

Carbon stocks play a crucial role in understanding carbon distribution and dynamic change patterns in ecosystems, as well as their capacity and absorption capacity of carbon sinks. Carbon stocks offer a scientific basis for predicting and assessing ecosystem responses to climate change, and the management and conservation of ecosystem carbon sinks (Ouyang & Lee et al., 2020; Richards et al., 2020). The examination of carbon stocks within different ecosystems can assess their potential and aid in optimizing ecosystem management, providing a scientific basis and guidance for ecology management and policymaking. Carbon stocks provide vital data for ecosystem carbon trade and markets, which facilitate the trading and transfer of ecosystem carbon sinks, thus promoting global carbon reduction and climate change response (Zhang et al., 2019; Ke et al., 2023). Exploring the response of carbon stock changes to climate change helps assess the impact of climate change on ecosystem carbon stocks, providing support for climate adaptation and adjustment (Wang et al., 2022; Paramesh et al., 2022). Carbon stock investigations are particularly valuable in assessing and managing the carbon sink capacity of ecosystems. Forests are the most significant carbon reservoir, and forest carbon sinks can efficiently reduce atmospheric $CO_2$ concentrations and mitigate climate change (Wang et al., 2020; Salimi et al., 2021). However, in regions or time periods where anthropogenic management and disturbance occur, interannual variability and dramatic land use change may transform forests into carbon sinks or sources (Dugan et al., 2021; Gogoi et al., 2022). Quantifying the spatial and temporal variability characteristics of regional carbon stocks can help explore the influencing factors and regulatory pathways of forest carbon source/sink functions (Launiainen et al., 2022).

The current trend in forest carbon stock monitoring and assessment is to integrate ground monitoring sample data and satellite observation data (Santoro et al., 2022; Lee et al., 2021). Ground-based monitoring, while producing high accuracy data, is time-consuming and not suitable for large-scale observation of forest areas (Liu et al., 2022; Teubner et al., 2019). In contrast, remote sensing inversion methods avoid the drawbacks of sample site monitoring, are more efficient, and are gradually improving in accuracy (Hamedianfar et al., 2022; Santoro et al., 2022). Spectral information-based methods can infer vegetation growth and type in ecosystems to estimate carbon stocks with high accuracy but are limited by image quality, making it difficult to explore deep non-linear relationships that would improve estimation accuracy (Lopatin et al., 2019). Conversely, structural information-based methods can directly measure biomass and carbon stocks, but are limited by remote sensing image resolution and coverage (Cuni-Sanchez et al., 2021; Sasmito et al., 2020). Model-based methods use carbon cycle models to simulate carbon sinks and ecosystem processes to estimate carbon stocks, taking into account differences and complexities between ecosystems, but require accurate ecological data and parameters (Zhao et al., 2019; Xiao et al., 2019). Machine learning-based methods can mine relationships between satellite images and carbon stocks quickly and efficiently (Safaei-Farouji et al., 2022; Li et al., 2022). Integrating sample monitoring and remote sensing data is an urgent problem that needs solving for constructing a universal and accurate regional forest carbon stock remote sensing monitoring model.

Medium- to high-resolution (10 ~ 30 m) optical data is currently among the most promising remote sensing data sources, and their long lifetime makes them suitable for continuously monitoring forest dynamics (Puliti et al., 2021). Most studies utilizing optical imagery have estimated forest biomass and stock volume, then calculated carbon stocks and sinks, with comparatively few studies performing direct estimation of carbon stocks from remote sensing imagery. Zhang et al. (2019) used the random forest algorithm to estimate and map 1 km of above-ground biomass in Chinese region by combining ground-based observations, MODIS, GLAS, and climate and topography data. Puliti et al. (2021) estimated total forest above-ground net change in a forest area in Norway (~1.4 million ha) using data from National Forest Inventory (NFI), Sentinel-2, and Landsat, with an RMSE of 45.5 Mg/ha. Chopping et al. (2022) estimated above-ground biomass in southwestern U.S. from 2000 to 2015 using multi-angle imaging spectra-radiometer (MISR) with an RMSE of 37.0 Mg/ha. Although high-resolution satellite images provide detailed spatial features and rich texture information, the existing carbon stock data products have relatively low resolution and few spectral bands. Proposing an effective algorithm to mine deep features is the key to accurately estimating carbon sinks.

Deep learning methods are widely used to extract deep features and achieve high accuracy (Lang et al., 2022; Lu et al., 2022; Santoro et al., 2022; Zhang et al., 2022). Current studies mainly use linear fitting and random forest methods for carbon stock estimation, with few utilizing deep learning methods. Style transfer methods utilizing transfer learning can reduce the domain shift of the original image, fix missing data and spectral discrepancies, and improve model generalizability for long time series. Since the proposal of Generative Adversarial Network (GAN) by Goodfellow et al. (2014, 2020), it had become the main method for image generation, including image restoration, style transfer, satellite image de-clouding, and noise reduction in the mainstream (Huang et al., 2019; Hui et al., 2021; Lateef et al., 2022; Pei et al., 2021). Different GAN methods, such as Conditional GAN (cGAN) and Pix2Pix, had been used for de-clouding (Bermudez et al., 2018; Turnes et al., 2020; Christovam et al., 2021), all generating reasonable cloud-free images but with spectral details that differ somewhat from real images. Transformer is a neural network model based on a self-attention mechanism mainly applied to natural language processing tasks and proposed by Google in 2017 (Vaswani et al., 2017). The Swin Transformer, proposed in 2021 based on window shifting, maps input features to different locations, reducing computational complexity and memory usage while achieving excellent performance on various vision tasks (Liu et al., 2021). Various Transformer-based models subsequently obtained state-of-the-arts (SOTA) in multiple fields (Jiang et al., 2023; Zeng et al., 2023; Zhong et al., 2023). Therefore, we use this as the backbone for performing remote sensing image style transfer and carbon stock estimation, resulting in advanced accuracy and enabling long-term monitoring.

In summary, our work involves converting carbon stock estimation into image translation, utilizing the Swin Transformer as a backbone network to develop the Swin-Pix2Pix algorithm with the UNet structure. We use GF-1 WFV images as data and consider the actual characteristics of Huize County region and forest to achieve high-resolution carbon stock estimation. Our results provide a theoretical basis for formulating regulations related to forest carbon sinks.

## 2 Material and methods

### 2.1 Study area

Huize County was situated in the northeastern part of Yunnan Province and the northwestern part of Qujing City. It was located at the junction of Yunnan, Sichuan, and Guizhou Provinces. The county occupied a land area of 5,889 km$^2$ as shown in Figure 1. The county was characterized by mountainous terrain that gradually decreases from west to east. The county's highest peak was 4,017 m above sea level, which was the highest peak in Qujing City. In contrast, its lowest point was at only 695 m above sea level, making it the lowest point in Qujing City. According to data from the Third National Land Survey with December 31, 2019, forest land covered 3,080.53 km$^2$ (~4,620,800 mu), with arboreal forest land accounting for 82.39% of the total area. The complexity of the county's physical geography presented a significant challenge, even though the abundance of forest resources provided ample samples for the study.

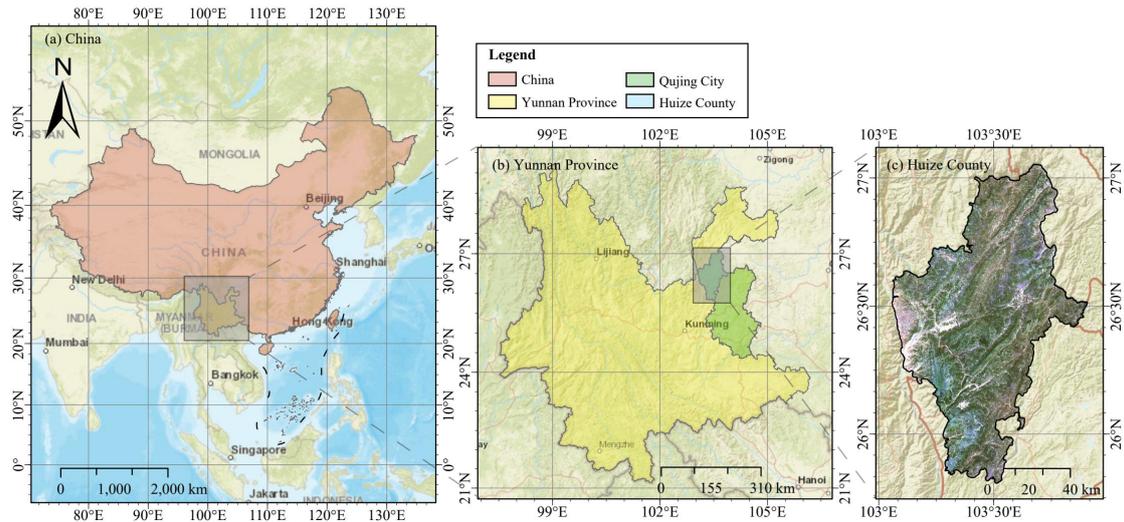

**Fig. 1** Study area. (a) is China, (b) is Yunnan Province, and (c) is Huize County.

## 2.2 Data sources

The study used the results of the Third National Land Survey (Third Survey) as the actual data, which were obtained from the Forestry and Grassland Bureau of Huize County. The data recorded more than 70 attributes such as accumulation, dominant tree species, small group area, tree species structure, etc., which can objectively represent the forest resources in the study area and can be useful for forest resources estimation. The survey period was October 8, 2017 to December 31, 2019, and had been fully completed in 2020. The survey had comprehensively refined and improved the basic data of China's land use and grasped the detailed and accurate status of it changes in natural resources.

The study utilized GF-1 WFV image data, which was the first satellite of China's high-resolution earth observation system launched in April 2013. The GF-1 WFV data had a spatial resolution of 16 m. To ensure temporal consistency, model training involved two scenes recorded on August 27, 2020, achieving full coverage of the study area. The data information of the images is shown in Table A1.

To achieve long-term monitoring, Landsat TM image data with a spatial resolution of 30 m was primarily used before 2013. To ensure data comparability, the Red, Green, Blue, and NIR bands were selected for both GF-1 WFV and Landsat TM images, and the data information is shown in Table 1. For Landsat images within the study area, the Path was 129, and the Rows were 41 and 42. Seamless mosaic tool in ENVI was used for stitching, and histogram matching was used for color correction.

An ALOS PALSAR DEM with a spatial resolution of 12.5 m was selected as the morphological reference. Advanced Land Observing Satellite (ALOS) aimed to contribute to the fields of mapping, precise regional land cover observation, disaster monitoring, and resource surveys, with DEM data as one of its products. From 2006 to 2011, it provided detailed all-day and all-season measurements. The data information is shown in Table A1.

**Table 1** Image band information of GF-1 WFV and Landsat TM.

| Description | GF-1 WFV | | | Landsat TM | | |
| --- | --- | --- | --- | --- | --- | --- |
| | Band Number | Wavelength (μm) | Resolution (m) | Band Number | Wavelength (μm) | Resolution (m) |
| Blue | Band 1 | 0.45~0.52 | 16 | Band 1 | 0.45~0.52 | 30 |
| Green | Band 2 | 0.52~0.59 | 16 | Band 2 | 0.52~0.60 | 30 |
| Red | Band 3 | 0.63~0.69 | 16 | Band 3 | 0.63~0.69 | 30 |
| NIR | Band 4 | 0.77~0.89 | 16 | Band 4 | 0.76~0.90 | 30 |

## 2.3 Methods

### 2.3.1 Data pre-processing

**Feature extraction.** The features extracted from the data comprise topographic, spectral, texture, and vegetation index features. A total of 50 bands were used, as shown in Table A2. The terrain features were obtained from DEM, while the rest were from GF-1 WFV. The extracted feature information is listed in Table 2. Topographic features included 11 bands such as slope and aspect, while spectral features consist of 4 bands of GF-1. Texture features were extracted using the gray-level co-occurrence matrix (GLCM), with 8 features comprising of mean and variance for each of the 4 bands of GF-1, totaling 32 feature bands. Finally, vegetation index features including the normalized difference vegetation index (NDVI), difference vegetation index (DVI), and ratio vegetation index (RVI) are calculated from Eqs. (1) ~ (3), where $\rho_{NIR}$ is the NIR band, and $\rho_{RED}$ is the red band.

$$NDVI = \frac{\rho_{NIR} - \rho_{RED}}{\rho_{NIR} + \rho_{RED}} \qquad (1)$$

$$RVI = \frac{\rho_{NIR}}{\rho_{RED}} \qquad (2)$$

$$DVI = \rho_{NIR} - \rho_{RED} \qquad (3)$$

**Table 2** Features information.

| Features | Spectral | Topographical | | | Vegetation Index | Texture | |
|---|---|---|---|---|---|---|---|
| **Indicators** | Band 1 | Slope | Plan Convexity | Maximum Curvature | NDVI | Mean | Dissimilarity |
| | Band 2 | Aspect | Longitudinal Convexity | RMS | DVI | Variance | Entropy |
| | Band 3 | Shaded Relief | Cross Sectional Convexity | Slope Percent | RVI | Homogeneity | Second Moment |
| | Band 4 | Profile Convexity | Minimum Curvature | - | - | Contrast | Correlation |
| **Total** | 4 bands | 11 bands | | | 3 bands | 32 bands | |

**Feature screening.** To avoid the limitation of a single correlation index, we selected bands that showed the highest correlation to carbon stock. We employed various measures, including the coefficients of Pearson, Spearman, and Kendall (Byakatonda et al., 2018), Cosine Similarity (Yin et al., 2022), and the distance of Euclidean, Manhattan, and Chebyshev (Zaitsev et al., 2017), to determine the correlation between each band and the measured value. From the average ranking of each band, we derived a comprehensive score. Correlation information for each band can be found in Table A3. Using this approach, the three bands that showed the highest correlation were 19th (GLCM-Mean_Band1), 35th (GLCM-Mean_Band3), and 27th (GLCM-Mean_Band2).

**Mask calculation.** The vegetation index can easily differentiate between vegetation and non-vegetation areas. In this paper, NDVI was selected to extract the spectral characteristics of forest land, which were regarded as the mask. The threshold calculation of the vegetation index is shown in Eq. (4).

$$M = \overline{M} - 2\sigma \tag{4}$$

Where, $M$ is the threshold, $\overline{M}$ is the average, and $\sigma$ is the standard deviation of vegetation index. With this formula, the threshold value in 2020 was 0.3951 (~0.40), where $\overline{M}$ was 0.6926 and $\sigma$ was 0.1487, and the NDVI was binarized, and the area above the threshold value was regarded as vegetation (set as 1) and the area below the threshold value was regarded as non-vegetation (set as 0). This extraction method was used to apply the mask calculation for the remaining years.

### 2.3.2 Swin Transformer Block (STB)

Transformer models (Vaswani et al., 2017) had been successful in the field of natural language processing (NLP) and had shown competitiveness beyond convolutional neural networks (CNN) in the image classification area (Dosovitskiy et al., 2020; Yuan et al., 2021). Transformers were designed for modeling global information and longer distance dependency relations, whereas CNNs were designed for modeling local information and were weaker in capturing global information. Transformers avoided the problem of bias towards particular examples present in CNNs. However, they were generally more complex than CNNs and not ideal for solving dense prediction tasks such as instance segmentation at the pixel level (Liu et al., 2018). Swin Transformer (Liu et al., 2021) addressed this problem with paned windows to reduce parameters for improved performance in many pixel-level vision tasks.

Swin Transformer used a hierarchical feature map building method similar to that used in CNN, which down-sampled images 4x, 8x, and 16x in the feature map. In the previous Vision Transformer (ViT), down-sampling was applied directly by 16x in the beginning, and the down-sampling ratio was maintained for subsequent feature maps. ViT produced a single low-resolution feature map due to the calculation of global self-attention, it had quadratic complexity in terms of input image size. With an increase in the depth of the network, the number of patches remained unchanged in the ViT model. In the case of Swin Transformer, the number of patches gradually reduced while the perceptual range of each patch increases. This design was intended to facilitate Swin Transformer's hierarchical construction and adapted to multi-scale visual tasks.

Swin Transformer adopted Windows Multi-Head Self-Attention (W-MSA), which partitioned the feature map into several disjointed areas or windows, specifically in 4x and 8x down-sampling ratios. Within each window, the model performed Multi-Head Self-Attention (MSA) only. In contrast to ViT, utilizing MSA directly on the global feature map, this technique mitigated computation, especially when the feature map was densely populated. However, it also compromised inter-window sharing. In response, Liu et al. (2021) proposed Shifted Windows Multi-Head Self-Attention (SW-MSA), which facilitated communication between adjacent windows. The network structure of Swin Transformer is illustrated in Figure 2(a).

The initial step involved inputting the image into the patch partition module for segmentation, where each patch comprised $4 \times 4$ adjacent pixels. Next, the patches were flattened in the channel direction. Assuming the input was an RGB image, each patch contained 16 pixels with three distinct values $(R, G, B)$. After flattening, the patch's shape changed from $[H, W, 3]$ to $[\frac{H}{4}, \frac{W}{4}, 48]$. Following this, the linear embedding layer conducted linear transformation on each pixel's channel data and converted 48 to $C$, thus modifying the image's shape from $[\frac{H}{4}, \frac{W}{4}, 48]$ to $[\frac{H}{4}, \frac{W}{4}, C]$. Notably, patch partition and linear embedding were actualized directly through a convolutional layer, sharing structural similarity with the embedding layer in ViT.

Subsequently, different feature maps of varying sizes were created through four stages. In stage one, aside from first using a linear embedding layer, the remaining three stages used a patch merging layer for down-sampling before repeating the Swin Transformer Block (STB) stacking. Two structures exist for the STB, as shown in Figure 2(b). The sole discrepancy between the structures was that one utilizes W-MSA, while the other utilized SW-MSA. Since these structures were paired, the STB was stacked an even number of times. The STB was consistent with the MLP structure in ViT.

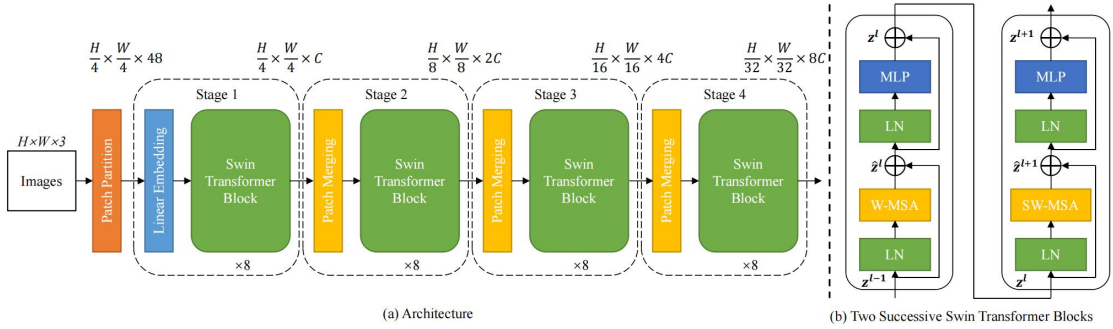

**Fig. 2** Swin Transformer (Swin-T) architecture. (a) is the overall architecture, and (b) is two successive Swin Transformer blocks.

### 2.3.3 Mask SUNet (MSUNet)

Swin U-Net (SUNet) first extracted shallow features using a $3 \times 3$ convolutional kernel and then main features using a U-Net structure, which replaced the original convolutional layers with STB to obtain high-level semantic information (Fan et al., 2021; Fan et al., 2022). SUNet comprised five STB layers and reconstructs the image using $3 \times 3$ convolutional kernels. SUNet replaced down-sampling with patch merging during encoding and up-sampling with dual up-sampling during decoding.

However, in our experiments, SUNet suffered from boundary blurring and noise issues. To address this problem, we introduced a mask to SUNet to filter background and boundary mixed image elements, which improved the clarity of the boundary in the estimation result. We also added a median filter to further reduce the noise, specifically speckle noise and salt-and-pepper noise. Mask SUNet architecture is shown in Figure 3.

**Patch Merging**. For the down-sampling module, we cascaded the input features of each $2 \times 2$ neighborhood block (Liu et al., 2021; Cao et al., 2021) and used a linear layer to obtain the output features with a specified number of channels. This step can be regarded as the initial stage of the convolution operation, which involved flattening the input feature map.

**Triple up-sample**. For up-sampling, the original Swin-UNet (Cao et al., 2021) used a transposed convolution-based block extension method in the up-sampling module. However, this method was prone to producing blocking artifacts. Fan et al. (2022) utilized dual up-sampling, which included two existing up-sampling methods, namely bilinear and pixel shuffle, and successfully prevented checkerboard artifacts. Nevertheless, dual up-sampling was not effective in defining boundaries in this particular task. We proposed the triple up-sample module by adding transposed convolution in addition to dual up-sample, which solved the problem of checkerboard artifacts and improves boundary definition.

**Median filter.** In estimation, noise was present around the mask in the output image, possibly due to the model's inability to estimate the boundary transition values well. The mask alone was not effective in filtering this noise. The median filter was effective in removing it, although it can lead to some image blurring. To address this problem, we found that the noise mainly consisted of high-intensity image elements. As a result, we removed only image elements above 240 intensity values. This method ensured that the image was not distorted while achieving effective noise removal.

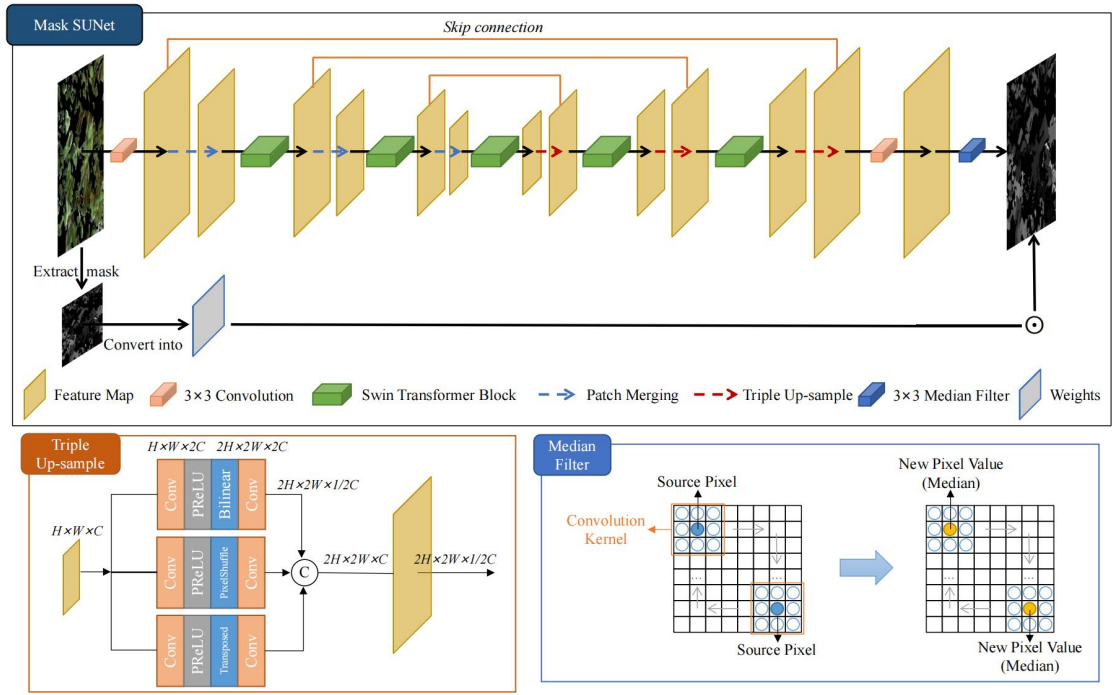

**Fig. 3** Mask SUNet architecture. The top panel represents Mask SUNet, while the bottom-left panel shows triple up-sample, and the bottom-right panel depicts the median filter.

### 2.3.4 MSwin-Pix2Pix

In the estimation, the generator used MSUNet and mask was used to filter the non-forest areas, which not only can make the estimation results with higher accuracy but also can increase the convergence speed of the model. The discriminator kept using the Patch GAN of Pix2Pix, and the loss function used $\mathsf{L}_{L1Smooth}$ and $\mathsf{L}_{L2}$. When reducing the domain differences between images captured at different times, the SUNet generator and $\mathsf{L}_{L1}$ were used, and mask was not needed for filtering.

Discriminators can severely affect the stability of adversarial training, and we kept Patch GAN unchanged. In experiments, we found that simply replacing the convolution with STB and increasing the number of model parameters made the training more stable under this baseline architecture. However, this pure Transformer architecture achieved little incremental benefit and increased the system overhead significantly, so we still used Patch GAN as the discriminator.

The discriminator of the original GAN was to output only one value (true or false), which was an evaluation of the whole image generated. The Patch GAN was designed in the form of a fully convolutional GAN. The true or false of each pixel in the $N \times N$ matrix represented the evaluation value of a small area (i.e., patch) in the original image, which was the application of the receptive field. Instead of measuring the whole image with a single value, the whole image was now evaluated using an $N \times N$ matrix. Patch discriminator (Isola et al., 2017) possessed limited receptive field and can be employed to specifically penalize the local structures. Experiments showed partial suppression of the blocking artifacts using a patch discriminator. MSwin-Pix2Pix architecture is shown in Figure 4.

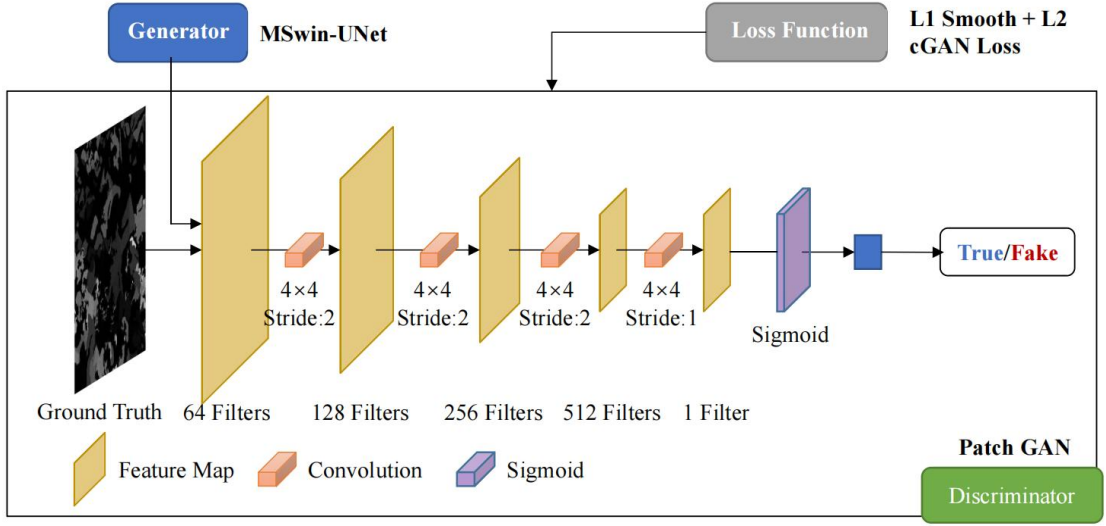

**Fig. 4** MSwin-Pix2Pix architecture.

### 2.3.5 Loss Function

(1) $\mathsf{L}_{L1}$ Loss

Isola et al. (2017) proposed the pix2pix cGAN using the $\mathsf{L}_{L1}$-distance, as shown in Eq. (5). $x$ was the ground truth, and $G(u)$ was the synthetic image using as condition the input data $u$. $\mathsf{L}_{L1}$ was also known as Mean Absolute Error (MAE) Loss. $\mathsf{L}_{L1}$ had a stable gradient for whatever input value, which did not lead to gradient explosion problem and had a more robust solution. However, it was not smooth at the zero point, where it was not derivable and converged more slowly. Generally, $\mathsf{L}_{L1}$ regularization created sparse features, where the weights of most useless features were set to zero, which had the effect of feature selection. The $\mathsf{L}_{L1}$ was also insensitive to noise and was more suitable for regression problems.

$$\mathsf{L}_{L1}(G) = \mathsf{E}_{x \sim P_{data}(x)} \left[ \| x - G(u) \| \right] \tag{1}$$

(2) $\mathsf{L}_{L2}$ Loss

Pathak et al. (2016) used the $\mathsf{L}_{L2}$-distance, as shown in Eq. (6). $\mathsf{L}_{L2}$ was also known as Mean Square Error (MSE) Loss. It was continuous and smooth at all points, easy to derive, and had a more stable solution. It was sensitive to outliers, and when the input value of the function was far from the true value, the corresponding value of the loss was large, then the gradient was large when solving using gradient descent, which might lead to gradient explosion. The $\mathsf{L}_{L2}$-distance was suitable for regression tasks with small numerical characteristics and low dimensionality of the problem.

$$\mathsf{L}_{L2}(G) = \mathsf{E}_{x \sim P_{data}(x)} \left[ (x - G(u))^2 \right] \tag{2}$$

(3) $\mathsf{L}_{L1Smooth}$ Loss

When the difference between the predicted and true values was small (the absolute value of the difference was less than 1), $\mathsf{L}_{L2}$ was used; when the difference was large, a translation of $\mathsf{L}_{L1}$ was used. $\mathsf{L}_{L1Smooth}$ was a combination of $\mathsf{L}_{L1}$ and $\mathsf{L}_{L2}$, leveraging the advantages of both approaches. $\mathsf{L}_{L1Smooth}$ modified the problem of unsmoothed zeros, and it was more robust to outliers than $\mathsf{L}_{L2}$. When using $\mathsf{L}_{L2}$, the gradient was smaller and the loss function was more rounded than $\mathsf{L}_{L1}$, which can converge faster. When using $\mathsf{L}_{L1}$, the gradient was small enough, more stable, and less prone to gradient explosion. In the regression task, it was more suitable when there were larger values in the features.

$$\mathsf{L}_{L1Smooth}(G) = \begin{cases} \mathsf{E}_{x \sim P_{data}(x)} \left[ (x - G(u))^2 \times 0.5 \right], if |x - G(u)| < 1 \\ \mathsf{E}_{x \sim P_{data}(x)} \left[ \|x - G(u)\| - 0.5 \right], \quad otherwise \end{cases} \tag{3}$$

(4) $\mathsf{L}_{cGAN}$ Loss

The original GAN contained a generator (G) and a discriminator (D), where G and D engaged in a minimax game. The G can learn to map a random noise vector (z) and observed image ($u$), to produce an output ($y$), as shown in Eq. (8). The objective function of cGAN is expressed in Eq. (9).

$$G : \{z, u\} \rightarrow y \tag{4}$$

$$\mathsf{L}_{cGAN}(G, D) = \mathsf{E}_{x \sim P_{data}(x)} \left[ \log D(x, u) \right] + \mathsf{E}_{z \sim P_{data}(z)} \left[ \log(1 - D(G(z, u))) \right] \tag{5}$$

(5) MSwin-Pix2Pix Loss

The loss function of Swin-Pix2Pix used for style transfer in this paper were $\mathsf{L}_{cGAN}$ and $\mathsf{L}_{L1}$, as shown in Eq. (10). While the losses of the MSwin-Pix2Pix model used for estimation were $\mathsf{L}_{cGAN}$, $\mathsf{L}_{L1Smooth}$ and $\mathsf{L}_{L2}$, as shown in Eq. (11), where $\lambda = 100$.

$$G^* = \arg \min_G \max_D \mathsf{L}_{cGAN}(G, D) + \lambda \mathsf{L}_{L1}(G) \tag{6}$$

$$G^* = \arg \min_G \max_D \mathsf{L}_{cGAN}(G, D) + \lambda \mathsf{L}_{L2}(G) + \lambda \mathsf{L}_{L1Smooth}(G) \tag{7}$$

### 2.3.6 Evaluation Metrics

Mean Absolute Error (MAE) was the mean of the absolute error between the predicted and true values as shown in Eq. (12). Mean Squared Error (MSE) was the mean of the absolute squared error between the predicted and true values as shown in Eq. (13). Structure Similarity Index Metrics (SSIM) was also a full-reference image quality evaluation metric, which measured image similarity in terms of luminance, contrast, and structure, respectively, as shown in Eq. (15). Where, $\hat{y}$ was the predicted value, $y$ was the true value, $\mu_{\hat{y}}$ and $\mu_y$ were the mean values of $\hat{y}$ and $y$, $\sigma_{\hat{y}}^2$ and $\sigma_y^2$ were the variances of $\hat{y}$ and $y$, respectively, and $\sigma_{\hat{y}y}$ was the covariance of $\hat{y}$ and $y$. $c_1 = (k_1 L)^2$, $c_2 = (k_2 L)^2$. Normally, $c_3 = c_2 / 2$.

$$MAE = \frac{1}{n}\sum_{i=1}^{n}|\hat{y}_i - y_i| \tag{1}$$

$$MSE = \frac{1}{n}\sum_{i=1}^{n}(\hat{y}_i - y_i)^2 \tag{2}$$

$$RMSE = \sqrt{\frac{1}{n}\sum_{i=1}^{n}(\hat{y}_i - y_i)^2} \tag{3}$$

$$SSIM(\hat{y}, y) = \frac{(2\mu_{\hat{y}}\mu_y + c_1)(2\sigma_{\hat{y}y} + c_2)}{(\mu_{\hat{y}}^2 + \mu_y^2 + c_1)(\sigma_{\hat{y}}^2 + \sigma_y^2 + c_2)} \tag{4}$$

# 3 Results

## 3.1 Ablation study for style transfer

The images at different temporal might have inter-domain differences due to the influence of sensors, lighting and others. To reduce such differences, we used a style transfer algorithm to reduce the distribution differences between the source and target domain and unified the spatial distribution of images at different temporal by style transfer across domains, which provided a basis for achieving the estimation the spatial distribution of carbon stocks. The selected images include 2005, 2013, 2015 and 2020, among which the 2005 image was Landsat TM image with a spatial resolution of 30 m and the rest images were GF-1 WFV with a spatial resolution of 16 m. The model selected for style transfer was Swin-Pix2Pix, and the results are shown in Figure 5. The last row was representative of de-clouding. Input $x$ was the image to be converted (i.e., 2005, 2013, 2015), the content reference, ground truth took the image in 2020 as the reference, i.e., the style reference, and output $G(x)$ was the converted image. It contained the content of input $x$ and ground truth style fusion image.

To enhance the quality of image stylization, the Swin-Pix2Pix algorithm used transformer structures, extracting global features into the generative network. This had the benefit of ensuring that network parameters did not intensify in one direction, allowing the parameter matrix to cover as many global features of the image as possible. Using global features to generate the image, the algorithm can increase diversity among generated samples, leading to improvements in image stylization quality. The improvement was particularly valuable in the de-clouding process, where thicker cloud regions contained more missing data requiring image restoration. Without rich global features, this can cause distorted restoration and had a severe impact on the next estimation step. Image stylization refered to the integration of stylized image attributes, including color and texture, with the original content image. It was in contrast to simple content image imitation. Therefore, the quality of the image stylization produced by the Swin-Pix2Pix algorithm was higher than that of Pix2Pix.

When using only $\mathsf{L}_{L1}$, the generated image was blurry. When using only $\mathsf{L}_{cGAN}$, the generated image was clear, but the color style was more different from the ground truth image. When using $\mathsf{L}_{L1} + \mathsf{L}_{cGAN}$, the generated images were clear again and retain more features of ground truth images. So, Swin-Pix2Pix used $\mathsf{L}_{L1} + \mathsf{L}_{cGAN}$.

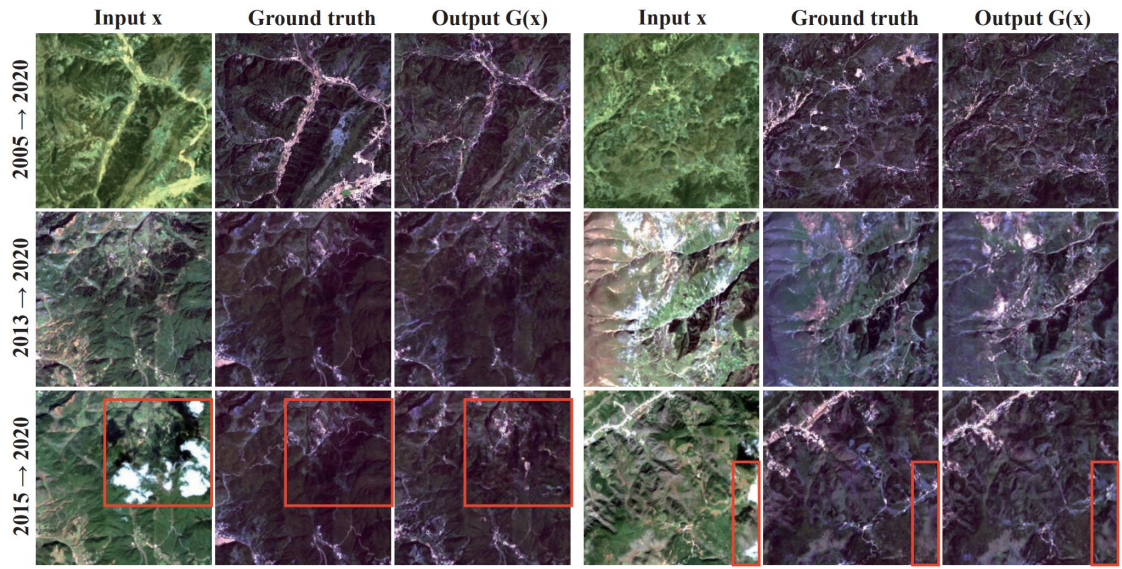

**Fig. 5** Style transfer results. Notes: Ground truth is the GF-1 WFV image in 2020, and the red box is the cloudy area.

**3.2 Ablation study for carbon stock estimation**

(1) Comparison of estimation results

The accuracy evaluation results are shown in Table 4, and the estimation results are shown in Figure 6. In addition to the proposed MSwin-Pix2Pix model, the comparison models included statistical models (i.e., OLS and GWR), machine learning models (i.e., RF and SVR), and deep learning models (i.e., CNN and Pix2Pix). To obtain stable evaluation, we utilized a five-fold cross-validation method with the dataset divided into training, validation, and test sets.

The $R^2$ of OLS, GWR, RF, and SVR were all below 0.5, and the SSIM were all below 0.4, indicating that the consistency between the estimation results and the measured values was poor, and both the numerical and spatial distributions were at a low level. Additionally, the RMSE and MAE values were found to be higher than average, suggesting that the estimation accuracy was low, and the error rate exceeded 100%. The deep learning models performed significantly better than the rest of the models, with CNN having the worst effect, followed by Pix2Pix, and MSwin-Pix2Pix having the best performance. Except for CNN, the $R^2$ of the remaining models were higher than 0.5, and the SSIM were higher than 0.65, indicating that the data consistency was better, but the spatial consistency was better than the numerical distribution, indicating that there was still room for improving the numerical accuracy. Among all models, MSwin-Pix2Pix had the highest estimation accuracy (MAE = 16.2891, RMSE = 29.3763, $R^2$ = 0.7105, SSIM = 0.7510).

Experimental results showed that the statistical model exhibited the poorest accuracy, followed by the traditional machine learning model, with the deep learning model performing the best. In particular, OLS and GWR models tended to extract linearly correlated features, resulting in insufficient regression analysis for tasks involving carbon stock estimation with texture features. Although RF and SVR models were commonly used in recent years for forest carbon stock and biomass estimation, they remained limited in terms of deeper feature extraction. CNN was a relatively basic deep learning model, and its estimation accuracy was significantly higher than that of the four models mentioned above. However, there was still room for improvement in the consistency of the estimation results. In this study, Swin Transformer was integrated into Pix2Pix to enhance its capability to extract global features, thereby improved model stability. The approach demonstrated the best performance among several models.

Median filter and mask were added to several deep learning models for the ablation experiments, as shown in Table 4. The addition of median filter alone improved the accuracy and filtered the noise surrounding the image effectively. Since the estimation results were required to be presented in blocks, a minor image blurring caused by the filtering was acceptable. Conversely, the performance improvement generated by the addition of the mask method was better than that produced by adding only the median filter approach. The outcome indicated that boundary detail feature extraction during model estimation still needed to be strengthened and suggested that noise levels in non-vegetative areas were lower than those in vegetation. Adding both median filter and mask modules to Swin-Pix2Pix significantly upgraded its performance more than other models. This improvement might be due to the fluctuations in the boundary transition region, the robust ability of the attention mechanism within the Transformer to extract global features, and the UNet structure's capacity to extract local features generating conflicts among the models. The median filter eliminated the anomalous detection points of local information, while the mask approach removed anomalous values from non-vegetation areas in the non-target process. Global feature extraction abilities of other models were limited; thus, this conflict did not occur. As a result, the median filter and mask methods proved more effective in Swin-Pix2Pix.

**Table 4** Compare of carbon stock estimation. Notes: OLS = Ordinary Least Squares; GWR = Geographically Weighted Regression; RF = Random Forest; SVR = Support

Vector Regress; CNN = Convolutional Neural Network. bold is the best, and underline is the second.

| Model | Mask | Median Filter | MAE | RMSE | R$^2$ | SSIM |
|---|---|---|---|---|---|---|
| OLS | - | - | 65.5938 | 104.9462 | 0.3858 | 0.2637 |
| GWR | - | - | 55.3973 | 93.3441 | 0.4370 | 0.3262 |
| RF | - | - | 52.9826 | 89.9311 | 0.4317 | 0.3653 |
| SVR | - | - | 50.2186 | 85.8489 | 0.4308 | 0.3588 |
| CNN | × | × | 31.5577 | 51.3224 | 0.4465 | 0.5731 |
| | × | ✓ | 31.0352 | 50.2659 | 0.4808 | 0.5692 |
| | ✓ | × | 31.7736 | 49.8384 | 0.4906 | 0.4863 |
| | ✓ | ✓ | 31.1258 | 49.3121 | 0.5077 | 0.5343 |
| Pix2Pix | × | × | 21.1221 | 38.6149 | 0.5203 | 0.6579 |
| | × | ✓ | 20.7252 | 38.0059 | 0.5331 | 0.6724 |
| | ✓ | × | 20.6300 | 37.9747 | 0.5368 | 0.6629 |
| | ✓ | ✓ | 20.5090 | 37.4499 | 0.5468 | 0.6698 |
| Swin-Pix2Pix | × | × | 18.9798 | 34.5187 | 0.6020 | 0.6986 |
| | × | ✓ | 18.3922 | 33.6495 | 0.6251 | <u>0.7167</u> |
| | ✓ | × | <u>18.0836</u> | <u>33.0752</u> | <u>0.6369</u> | 0.7011 |
| MSwin-Pix2Pix | ✓ | ✓ | **16.2891** | **29.3763** | **0.7105** | **0.7510** |

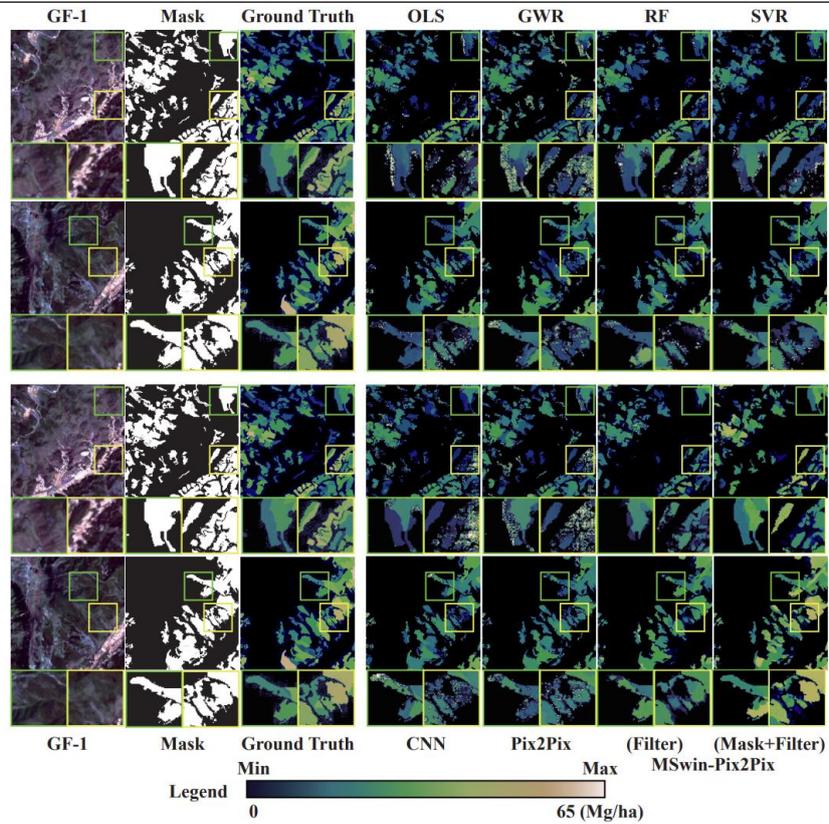

**Fig. 6** Carbon stock estimation results.

(2) Comparison of loss function

$L_{L1}$ produced a fixed gradient for any input value, which prevented gradient spikes and enhances its robustness. However, it was folded at the center point, cannot be derived, and might prompt gradient oscillation or even disappearance. $L_{L2}$, being continuous and smooth, can be derived at every point, gave a more stable solution, and did not cause gradient oscillation. However, it was not robust and may lead to gradient explosion problems due to too large input values. The $L_{L1Smooth}$ was more robust to outliers in comparison with $L_{L2}$, indicating it was insensitive to outliers distant from the central point. Moreover, $L_{L1Smooth}$ can regulate the gradient's magnitude to prevent it from escaping during the training process. If $L_{L2}$ was used in Pix2Pix, the average of all confidence outputs was computed, and this might cause blurry images. Additionally, $L_{L2}$ was highly sensitive to noise and using $L_{L1}$ provided less attention to samples with large prediction differences, making it a more intuitive approach.

The measured carbon stock data used in this paper were monitored in small-plot, non-sample point data, and were distributed in blocks. In MSwin-Pix2Pix for carbon stock estimation, mask was added to the model to filter the boundary non-vegetation image elements, and median filter was added to smooth the data. On this basis, the effect of image blurring can be ignored, and the estimation process using $L_{L2}$ can obtain more accurate and stable estimation results than $L_{L1}$; the robustness of $L_{L1Smooth}$ to noise compensated the sensitivity of $L_{L2}$ to outliers. Therefore, the optimal performance can be obtained by using $\mathsf{L}_{L1Smooth}$ and $\mathsf{L}_{L2}$ at the same time. While using $\mathsf{L}_{L1}$, $\mathsf{L}_{L1Smooth}$ and $\mathsf{L}_{L2}$ simultaneously decreased the effect, which might be due to the conflict between the zero-integrable nature of $\mathsf{L}_{L1}$ and $\mathsf{L}_{L1Smooth}$.

**Table 3** Compare of loss function. Notes: All experiments included $\mathsf{L}_{cGAN}$. Bold is the best, and underline is the second.

| Loss | $\mathsf{L}_{L1}$ | $\mathsf{L}_{L2}$ | $\mathsf{L}_{L1Smooth}$ | MAE | RMSE | R² | SSIM |
|---|---|---|---|---|---|---|---|
| $\mathsf{L}_{L1}$ | ✓ | ✗ | ✗ | 17.4877 | 31.6513 | 0.6634 | 0.6902 |
| $\mathsf{L}_{L2}$ | ✗ | ✓ | ✗ | 17.6498 | <u>30.0438</u> | 0.4197 | 0.7301 |
| $\mathsf{L}_{L1Smooth}$ | ✗ | ✗ | ✓ | 18.6862 | 31.5085 | 0.3609 | 0.6985 |
| $\mathsf{L}_{L1} + \mathsf{L}_{L2}$ | ✓ | ✓ | ✗ | 18.7327 | 31.0134 | 0.3916 | 0.7238 |
| $\mathsf{L}_{L1} + \mathsf{L}_{L1Smooth}$ | ✓ | ✗ | ✓ | 17.2201 | 31.0997 | 0.6747 | 0.7113 |
| $\mathsf{L}_{L2} + \mathsf{L}_{L1Smooth}$ | ✗ | ✓ | ✓ | **16.2891** | **29.3763** | **0.7105** | **0.7510** |
| $\mathsf{L}_{L2} + \mathsf{L}_{L1} + \mathsf{L}_{L1Smooth}$ | ✓ | ✓ | ✓ | <u>16.8845</u> | 30.5144 | <u>0.6878</u> | <u>0.7408</u> |

### 3.3 Spatial and temporal variation characteristics of carbon stocks

Over the 15-year period from 2005 to 2020, the area with increasing carbon stock covered 2,593.76 km² (about 3,890,600 mu) which represented 44.04% of the total area studied. In contrast, the carbon stock decreased in an area of 601.56 km² (about 902,300 mu) accounting for 10.22% of the total area while the remaining 45.74% had no significant change in their carbon stock. The areas with invariant and increasing carbon stock are shown in Figure 7, and the spatial and temporal distribution characteristics of carbon stock for four periods from 2005 to 2020 are shown in Figure A1. The carbon stock of Huize County had an overall increasing trend. 2,293.93 km$^2$ (about 3,440,900 mu) of forested land was available in 2005, and 2,645.02 km$^2$ (about 3,967,500 mu) in 2020, accounting for 44.15% of the total area of the county. The standing volume of live trees in 2005 was 10,548,990 m$^3$, of which the volume of the forest was 10,286,700 m$^3$, accounting for 97.51% of the total standing volume. The forest coverage rate was 39.9%, which increased to 50.38% in 2020.

The ecological environment had improved significantly, laying a good ecological foundation for the overall social and economic development of the county. The forestry industry had developed comprehensively, and as of 2020, the area of nature reserves had reached 169.73 km$^2$ (254,600 mu), and ecological and species protection had been strengthened. The pure forests of *Pinus yunnanensis* Franch and *Pinus armandii* Franch in the county accounted for 89.73% of the tree forest area. The forest resources in Huize County were characterized by the following: greater abundance towards the east while decreasing towards the west, sparser distribution in the north and south, predominantly pure forests as opposed to mixed forests, predominantly needleleaf forests as opposed to broadleaf forests, greater representation of planted forests than natural forests, and a relatively homogeneous tree species structure.

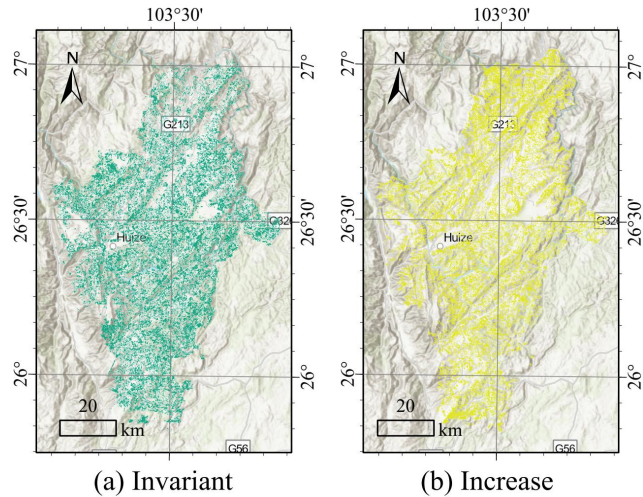

(a) Invariant          (b) Increase

**Fig. 7** Carbon stock temporal and spatial variation characteristics.

## 4 Discussion

### 4.1 Reduce domain shift

During long-term and large-scale estimation, satellite images captured at different times and locations can experience distortion from sensor and illumination effects. The matched histogram and relative radiation correction were currently the main solutions, although they each have specific limitations (Zhang et al., 2023; Roy et al., 2008). The matched histogram enhances image contrast by rescaling pixel values and altering hue. Nonetheless, this method was vulnerable to noise, cannot handle nonlinear transformations, and was dependent on a customized target histogram, which can compromise image quality if not set properly. In contrast, the relative radiation correction method can eliminate shadows to improve the overall image brightness while maintaining image detail, making it applicable across diverse lighting conditions. Cloudiness was a significant limitation to data availability in highland regions for several years consecutively, as demonstrated in existing research (Christovam et al., 2022). In this paper, we employed the highly effective transfer learning-based image style transfer model to address disparities caused by variations in temporal, sensors, and illumination conditions.

Our approach involved merging the Pix2Pix style transfer model with the Swin Transformer backbone network to extract global features. This alignment mechanism served to minimize the domain offset and improve image fusion of content and style. The SUNet had better feature extraction than UNet, largely due to the superior performance of Swin Transformer. As a result, the image style transfer and generation quality of Swin-Pix2Pix was superior to that of Pix2Pix, regarding both texture detail and spectral distribution features.

## 4.2 Carbon stock estimation

Huize County boasted substantial forest resources that provide a firm foundation for this study. The complex physical geography arised from altitude differences results in significant challenges for accurate carbon stock estimation. Current research in carbon stock estimation through satellite imagery was divided into LiDAR and optical images. LiDAR offered higher accuracy than most multispectral imagery, at the expense of difficult data acquisition and scalability limitations. Optical imagery compensated for this deficiency, thanks to the high spatiotemporal resolution advantage of the GF-1 satellite. Moreover, this paper addressed the issue of cloud and rain interference, which was resolved using the image transfer method. Thus, high-accuracy optical-image estimation posed a potential challenge for future carbon stock estimation research.

Multi-temporal carbon stock estimation studied commonly employ limited machine learning models such as OLS, RF, and SVR (Eckert, 2012; Yadav & Nandy, 2015; Zhang et al., 2018). The relationship between spectral and textural characteristics and biomass, accumulation and carbon storage were not a simple linear relationship. To address this limitation, we added a Transformer backbone network that employs deep learning theory, an attention mechanism to extract global features, and a median filter module to extract local features. Moreover, non-vegetated areas were masked via filtering, and the proposed MSwin-Pix2Pix model can extract deeper features. The model improved estimation accuracy (RMSE = 29.3763), which was comparable to that of LiDAR estimation (RMSE = 25.64) (Cao et al., 2016) and better than multi-source and multi-temporal imagery with coarse resolution (1 km) (RMSE ≈ 30) (Chen et al., 2023). Although Cao et al. (2016) and Chen et al. (2023) were biomass estimates, they were comparable to carbon stocks due to their strong correlation. The high-resolution multi-temporal images had the potential to estimate carbon stocks, suggesting accurate estimates.

**5 Conclusion**

This paper took Huize County, Qujing City, Yunnan Province, China as the study area, used GF-1 WFV and Landsat TM images as data, and proposed the Swin-Pix2Pix model for domain transfer of multi-temporal images based on the style transfer model Pix2Pix with deep learning, and introduced Swin Transformer to extract global features through attention mechanism and de-clouding processing. Additionally, we applied a mask and median filter to construct the MSwin-Pix2Pix model for precise estimation of carbon stocks. Our research provided a foundation for accurate estimates of regional carbon sinks. The main conclusions were:

(1) To reduce the domain offset, the Swin-Pix2Pix was used to decrease the distance between the source and target domain. Aligning the spatial distribution of different temporal by style transfer across domains, this method resolved the inter-domain differences caused by various factors like temporal, sensors and lighting. With enhanced image de-clouding, this approach outperforms Pix2Pix, provided a data basis for the long-term estimation of carbon stock spatial distribution.

(2) Deep learning models can extract deep features for carbon stock estimation, outperforming most other models. We incorporated the Swin Transformer to Pix2Pix, adding the ability to extract global features and enhance model stability. The median filter module had been implemented to eliminate detection anomalies using local information, while the mask module removed non-target regions for higher accuracy. The proposed MSwin-Pix2Pix outperforms other models in terms of estimation accuracy (MAE = 16.2891, RMSE = 29.3763, $R^2$ = 0.7105, SSIM = 0.7510).

(3) Carbon stock spatial and temporal characteristics showed that the region with increasing carbon stock coverage was 2,593.76 km$^2$, accounting for 44.04% of the total area, while the areas with a decrease were 601.56 km$^2$, accounting for 10.22%, and the area where it remained stagnant was 45.74%. Forest coverage increased over time, from 39.9% in 2005 to 50.38% in 2020, and the overall trend for carbon storage was increasing. Environmental improvements had been significant, laying a good ecological foundation for the overall socio-economic development of the region.

**Acknowledgements**

This research was funded by National Key R&D Program of China [2021YFE0117300].

**Author contributions**

Jinnian Wang supervised and organized the project. Zhenyu Yu developed the code and wrote the manuscript. Yuqing Xie provided the in-situ data. All authors revised the manuscript.

**Competing interests**

The authors declare no competing interests.

**References**

Bermudez, J. D., Happ, P. N., Oliveira, D. A. B., & Feitosa, R. Q. (2018). Sar to Optical Image Synthesis for Cloud Removal with Generative Adversarial Networks. *ISPRS Annals of the Photogrammetry, Remote Sensing and Spatial Information Sciences*, *IV–1*, 5–11. https://doi.org/10.5194/isprs-annals-IV-1-5-2018

Byakatonda, J., Parida, B. P., Moalafhi, D. B., & Kenabatho, P. K. (2018). Analysis of long term drought severity characteristics and trends across semiarid

Botswana using two drought indices. *Atmospheric Research*, *213*, 492–508. https://doi.org/10.1016/j.atmosres.2018.07.002

Cao, H., Wang, Y., Chen, J., Jiang, D., Zhang, X., Tian, Q., & Wang, M. (2023). *Swin-Unet: Unet-like Pure Transformer for Medical Image Segmentation*. In Computer Vision–ECCV 2022 Workshops, 205-218. http://arxiv.org/abs/2105.05537

Cao, L., Coops, N. C., Innes, J. L., Sheppard, S. R. J., Fu, L., Ruan, H., & She, G. (2016). Estimation of forest biomass dynamics in subtropical forests using multi-temporal airborne LiDAR data. *Remote Sensing of Environment*, *178*, 158–171. https://doi.org/10.1016/j.rse.2016.03.012

Chen, Y., Feng, X., Fu, B., Ma, H., Zohner, C. M., Crowther, T. W., Huang, Y., Wu, X., & Wei, F. (2023). Maps with 1 km resolution reveal increases in above- and belowground forest biomass carbon pools in China over the past 20 years. *Earth System Science Data*, *15*(2), 897–910. https://doi.org/10.5194/essd-15-897-2023

Chopping, M., Wang, Z., Schaaf, C., Bull, M. A., & Duchesne, R. R. (2022). Forest aboveground biomass in the southwestern United States from a MISR multi-angle index, 2000–2015. *Remote Sensing of Environment*, *275*, 112964. https://doi.org/10.1016/j.rse.2022.112964

Christovam, L. E., Shimabukuro, M. H., Galo, M. de L. B. T., & Honkavaara, E. (2022). Pix2pix Conditional Generative Adversarial Network with MLP Loss Function for Cloud Removal in a Cropland Time Series. *Remote Sensing*, *14*(1),

Article 1. https://doi.org/10.3390/rs14010144

Dosovitskiy, A., Beyer, L., Kolesnikov, A., Weissenborn, D., Zhai, X., Unterthiner, T., Dehghani, M., Minderer, M., Heigold, G., Gelly, S., Uszkoreit, J., & Houlsby, N. (2020). *An Image is Worth 16x16 Words: Transformers for Image Recognition at Scale* (arXiv:2010.11929; Version 1). arXiv. https://doi.org/10.48550/arXiv.2010.11929

Dugan, A. J., Lichstein, J. W., Steele, A., Metsaranta, J. M., Bick, S., & Hollinger, D. Y. (2021). Opportunities for forest sector emissions reductions: A state-level analysis. *Ecological Applications*, *31*(5). https://doi.org/10.1002/eap.2327

Eckert, S. (2012). Improved Forest Biomass and Carbon Estimations Using Texture Measures from WorldView-2 Satellite Data. *Remote Sensing*, *4*(4), Article 4. https://doi.org/10.3390/rs4040810

Fan, B., Dai, Y., & He, M. (2021). Sunet: symmetric undistortion network for rolling shutter correction. *In Proceedings of the IEEE/CVF International Conference on Computer Vision* (pp. 4541-4550). https://doi.org/10.48550/arXiv.2108.04775

Fan, C. M., Liu, T. J., & Liu, K. H. (2022). SUNet: Swin Transformer UNet for Image Denoising. *2022 IEEE International Symposium on Circuits and Systems (ISCAS)*, 2333–2337. https://doi.org/10.1109/ISCAS48785.2022.9937486

Gao, W., Shen, F., Tan, K., Zhang, W., Liu, Q., Lam, N. S., & Ge, J. (2021). Monitoring terrain elevation of intertidal wetlands by utilising the spatial-temporal fusion of multi-source satellite data: A case study in the

Yangtze (Changjiang) Estuary. *Geomorphology*, *383*, 107683. https://doi.org/10.1016/j.geomorph.2021.107683

Gogoi, A., Ahirwal, J., & Sahoo, U. K. (2022). Evaluation of ecosystem carbon storage in major forest types of Eastern Himalaya: Implications for carbon sink management. *Journal of Environmental Management*, *302*, 113972. https://doi.org/10.1016/j.jenvman.2021.113972

Goodfellow, I., Pouget-Abadie, J., Mirza, M., Xu, B., Warde-Farley, D., Ozair, S., ... & Bengio, Y. (2014). Generative adversarial nets in advances in neural information processing systems (NIPS). Curran Associates, Inc. Red Hook, NY, USA, 2672-2680.

Goodfellow, I. J., Pouget-Abadie, J., Mirza, M., Xu, B., Warde-Farley, D., Ozair, S., & Bengio, Y. (2014). Generative Adversarial Networks. *Advances in neural information processing systems*, 1–9. arXiv:1406.2661. https://doi.org/10.48550/arXiv.1406.2661

Goodfellow, I., Pouget-Abadie, J., Mirza, M., Xu, B., Warde-Farley, D., Ozair, S., ... & Bengio, Y. (2020). Generative adversarial networks. *Communications of the ACM*, *63*(11), 139-144. https://doi.org/10.1145/3422622

Hamedianfar, A., Mohamedou, C., Kangas, A., & Vauhkonen, J. (2022). Deep learning for forest inventory and planning: A critical review on the remote sensing approaches so far and prospects for further applications. *Forestry: An International Journal of Forest Research*, *95*(4), 451–465.

https://doi.org/10.1093/forestry/cpac002

Huang, Y., Lu, Z., Shao, Z., Ran, M., Zhou, J., Fang, L., & Zhang, Y. (2019). Simultaneous denoising and super-resolution of optical coherence tomography images based on generative adversarial network. *Optics Express*, *27*(9), 12289. https://doi.org/10.1364/OE.27.012289

Hui, Z., Li, J., Gao, X., & Wang, X. (2021). Progressive perception-oriented network for single image super-resolution. *Information Sciences*, *546*, 769–786. https://doi.org/10.1016/j.ins.2020.08.114

Isola, P., Zhu, J. Y., Zhou, T., & Efros, A. A. (2018). Image-to-image translation with conditional adversarial networks. *In Proceedings of the IEEE conference on computer vision and pattern recognition* (pp. 1125-1134). https://doi.org/10.48550/arXiv.1611.07004

Jiang, P., Deng, F., Wang, X., Shuai, P., Luo, W., & Tang, Y. (2023). Seismic First Break Picking Through Swin Transformer Feature Extraction. *IEEE Geoscience and Remote Sensing Letters*, *20*, 1–5. https://doi.org/10.1109/LGRS.2023.3248233

Lang, N., Kalischek, N., Armston, J., Schindler, K., Dubayah, R., & Wegner, J. D. (2022). Global canopy height regression and uncertainty estimation from GEDI LIDAR waveforms with deep ensembles. *Remote Sensing of Environment*, *268*, 112760. https://doi.org/10.1016/j.rse.2021.112760

Lateef, F., Kas, M., & Ruichek, Y. (2022). Saliency Heat-Map as Visual Attention for Autonomous Driving Using Generative Adversarial Network (GAN). *IEEE Transactions on Intelligent Transportation Systems*, *23*(6), 5360–5373. https://doi.org/10.1109/TITS.2021.3053178

Launiainen, S., Katul, G. G., Leppä, K., Kolari, P., Aslan, T., Grönholm, T., Korhonen, L., Mammarella, I., & Vesala, T. (2022). Does growing atmospheric $CO_2$ explain increasing carbon sink in a boreal coniferous forest? *Global Change Biology*, *28*(9), 2910–2929. https://doi.org/10.1111/gcb.16117

Lee, J., Kim, B., Noh, J., Lee, C., Kwon, I., Kwon, B.-O., Ryu, J., Park, J., Hong, S., Lee, S., Kim, S.-G., Son, S., Yoon, H. J., Yim, J., Nam, J., Choi, K., & Khim, J. S. (2021). The first national scale evaluation of organic carbon stocks and sequestration rates of coastal sediments along the West Sea, South Sea, and East Sea of South Korea. *Science of The Total Environment*, *793*, 148568. https://doi.org/10.1016/j.scitotenv.2021.148568

Liu, S., Qi, L., Qin, H., Shi, J., & Jia, J. (2018). Path aggregation network for instance segmentation. *Proceedings of the IEEE Conference on Computer Vision and Pattern Recognition*, 8759–8768.

Liu, Z., Lin, Y., Cao, Y., Hu, H., Wei, Y., Zhang, Z., Lin, S., & Guo, B. (2021). *Swin Transformer: Hierarchical Vision Transformer using Shifted Windows* (arXiv:2103.14030). arXiv. http://arxiv.org/abs/2103.14030

Liu, Z., Zhao, F., Liu, X., Yu, Q., Wang, Y., Peng, X., Cai, H., & Lu, X. (2022). Direct estimation of photosynthetic CO2 assimilation from solar-induced chlorophyll fluorescence (SIF). *Remote Sensing of Environment*, *271*, 112893. https://doi.org/10.1016/j.rse.2022.112893

Lu, W., Tao, C., Li, H., Qi, J., & Li, Y. (2022). A unified deep learning framework for urban functional zone extraction based on multi-source heterogeneous data. *Remote Sensing of Environment*, *270*, 112830. https://doi.org/10.1016/j.rse.2021.112830

Noa Turnes, J., Castro, J. D. B., Torres, D. L., Vega, P. J. S., Feitosa, R. Q., & Happ, P. N. (2022). Atrous cGAN for SAR to Optical Image Translation. *IEEE Geoscience and Remote Sensing Letters*, *19*, 1–5. https://doi.org/10.1109/LGRS.2020.3031199

Pathak, D., Krähenbühl, P., Donahue, J., Darrell, T., & Efros, A. A. (2016). Context Encoders: Feature Learning by Inpainting. *2016 IEEE Conference on Computer Vision and Pattern Recognition (CVPR)*, 2536–2544. https://doi.org/10.1109/CVPR.2016.278

Pei, Z., Jin, M., Zhang, Y., Ma, M., & Yang, Y.-H. (2021). All-in-focus synthetic aperture imaging using generative adversarial network-based semantic inpainting. *Pattern Recognition*, *111*, 107669. https://doi.org/10.1016/j.patcog.2020.107669

Puliti, S., Breidenbach, J., Schumacher, J., Hauglin, M., Klingenberg, T. F., & Astrup, R. (2021). Above-ground biomass change estimation using national forest inventory data with Sentinel-2 and Landsat. *Remote Sensing of Environment*, *265*, 112644. https://doi.org/10.1016/j.rse.2021.112644

Roy, D. P., Ju, J., Lewis, P., Schaaf, C., Gao, F., Hansen, M., & Lindquist, E. (2008). Multi-temporal MODIS–Landsat data fusion for relative radiometric normalization, gap filling, and prediction of Landsat data. *Remote Sensing of Environment*, *112*(6), 3112–3130. https://doi.org/10.1016/j.rse.2008.03.009

Salimi, S., Berggren, M., & Scholz, M. (2021). Response of the peatland carbon dioxide sink function to future climate change scenarios and water level management. *Global Change Biology*, *27*(20), 5154–5168. https://doi.org/10.1111/gcb.15753

Santoro, M., Cartus, O., & Fransson, J. E. S. (2022a). Dynamics of the Swedish forest carbon pool between 2010 and 2015 estimated from satellite L-band SAR observations. *Remote Sensing of Environment*, *270*, 112846. https://doi.org/10.1016/j.rse.2021.112846

Santoro, M., Cartus, O., & Fransson, J. E. S. (2022b). Dynamics of the Swedish forest carbon pool between 2010 and 2015 estimated from satellite L-band SAR observations. *Remote Sensing of Environment*, *270*, 112846. https://doi.org/10.1016/j.rse.2021.112846

Teubner, I. E., Forkel, M., Camps-Valls, G., Jung, M., Miralles, D. G., Tramontana, G., van der Schalie, R., Vreugdenhil, M., Mösinger, L., & Dorigo, W. A. (2019). A carbon sink-driven approach to estimate gross primary production from microwave satellite observations. *Remote Sensing of Environment*, *229*, 100–113. https://doi.org/10.1016/j.rse.2019.04.022

Vaswani, A., Shazeer, N., Parmar, N., Uszkoreit, J., Jones, L., Gomez, A. N., ... & Polosukhin, I. (2017). Attention is all you need. *Advances in neural information processing systems*, *30*. https://doi.org/10.48550/arXiv.1706.03762

Wang, J., Feng, L., Palmer, P. I., Liu, Y., Fang, S., Bösch, H., O'Dell, C. W., Tang, X., Yang, D., & Liu, L. (2020). Large Chinese land carbon sink estimated from atmospheric carbon dioxide data. *Nature*, *586*(7831), 720–723. https://doi.org/10.1038/s41586-020-2849-9

Yadav, B. K. V., & Nandy, S. (2015). Mapping aboveground woody biomass using forest inventory, remote sensing and geostatistical techniques. *Environmental Monitoring and Assessment*, *187*(5), 308. https://doi.org/10.1007/s10661-015-4551-1

Yin, J., & Sun, S. (2022). Incomplete multi-view clustering with cosine similarity. *Pattern Recognition*, *123*, 108371. https://doi.org/10.1016/j.patcog.2021.108371

Yuan, L., Chen, Y., Wang, T., Yu, W., Shi, Y., Jiang, Z., Tay, F. E. H., Feng, J., & Yan, S. (2021). Tokens-to-Token ViT: Training Vision Transformers from Scratch on ImageNet. *2021 IEEE/CVF International Conference on Computer Vision (ICCV)*, 538–547. https://doi.org/10.1109/ICCV48922.2021.00060

Zaitsev, D. A. (2017). A generalized neighborhood for cellular automata. *Theoretical Computer Science*, *666*, 21–35. https://doi.org/10.1016/j.tcs.2016.11.002

Zeng, Y., Yang, X., Pan, L., Zhu, W., Wang, D., Zhao, Z., Liu, J., Sun, C., & Zhou, C. (2023). Fish school feeding behavior quantification using acoustic signal and improved Swin Transformer. *Computers and Electronics in Agriculture*, *204*, 107580. https://doi.org/10.1016/j.compag.2022.107580

Zhang, C., Denka, S., Cooper, H., & Mishra, D. R. (2018). Quantification of sawgrass marsh aboveground biomass in the coastal Everglades using object-based ensemble analysis and Landsat data. *Remote Sensing of Environment*, *204*, 366–379. https://doi.org/10.1016/j.rse.2017.10.018

Zhang, L., Su, G., Yin, J., Li, Y., Lin, Q., Zhang, X., & Shao, L. (2022). Bioinspired Scene Classification by Deep Active Learning with Remote Sensing Applications. *IEEE Transactions on Cybernetics*, *52*(7), 5682–5694. https://doi.org/10.1109/TCYB.2020.2981480

Zhang, R., Zhou, X., Ouyang, Z., Avitabile, V., Qi, J., Chen, J., & Giannico, V. (2019). Estimating aboveground biomass in subtropical forests of China by integrating multisource remote sensing and ground data. *Remote Sensing of Environment*, *232*, 111341. https://doi.org/10.1016/j.rse.2019.111341

Zhong, B., Wei, T., Luo, X., Du, B., Hu, L., Ao, K., Yang, A., & Wu, J. (2023). Multi-Swin Mask Transformer for Instance Segmentation of Agricultural Field Extraction. *Remote Sensing*, *15*(3), Article 3. https://doi.org/10.3390/rs15030549